\documentclass[10pt,twocolumn,letterpaper]{article}

\usepackage{iccv}
\usepackage{times}
\usepackage{epsfig}
\usepackage{graphicx}
\usepackage{amsmath}
\usepackage{amssymb}
\usepackage{comment}

\usepackage[breaklinks=true,bookmarks=false]{hyperref}
\iccvfinalcopy 

\usepackage[utf8]{inputenc} 
\usepackage[T1]{fontenc}    
\usepackage{hyperref}       
\usepackage{url}            
\usepackage{booktabs}       
\usepackage{amsfonts}       

\usepackage{nicefrac}       
\usepackage{microtype}      
\usepackage{adjustbox}
\usepackage{booktabs}
\usepackage{multirow}
\usepackage{adjustbox}
\usepackage{xcolor}
\usepackage{graphics,psfrag}
\usepackage{diagbox} 
\usepackage{subfigure}
\usepackage{bbding}
\usepackage{amssymb}

\usepackage[switch]{lineno}
\usepackage{xcolor,soul,colortbl}
\newcommand{\norm}[1]{\lVert#1\rVert}
\newcommand{\jun}[1]{\textcolor{magenta}{Jun: #1}}
\newcommand{\ron}{\romannumeral}
\newcommand{\widthscalefive}{0.1}
\newcolumntype{C}[1]{>{\centering\arraybackslash}m{#1}}
\newcolumntype{R}[1]{>{\raggedleft\arraybackslash}m{#1}}
\newcolumntype{P}[1]{>{\raggedright\arraybackslash}p{#1}}
\newcolumntype{M}[1]{>{\centering\arraybackslash}m{#1}}
\def\etal{\emph{et al.}}
\usepackage{floatrow}

\newfloatcommand{figurebox}{figure}[\nocapbeside][\dimexpr(\textwidth-\columnsep)/2\relax]
\newfloatcommand{tablebox}{table}[\nocapbeside][\dimexpr(\textwidth-\columnsep)/2\relax]
\def\iccvPaperID{3033} 
\def\httilde{\mbox{\tt\raisebox{-.5ex}{\symbol{126}}}}

\ificcvfinal\fi

\begin{document}
	
	\title{Towards Discriminative Representation Learning  for Unsupervised Person Re-identification}
	
	\author{%
		Takashi Isobe$^{1,2}$, Dong Li$^1$, Lu Tian$^1$, \\ Weihua Chen$^3$, Yi Shan$^1$,  Shengjin Wang$^2$\thanks{Corresponding author}\\
		{$^1$Xilinx Inc., Beijing, China.}\\
		{$^2$Department of Electronic Engineering, Tsinghua University}\\
		{$^3$Machine Intelligence Technology Lab, Alibaba Group}\\
		{\texttt{\small{\{dongl, lutian, yishan\}}@xilinx.com}\hspace{0.5cm}}
		{\texttt{\small{jbj18@mails.tsinghua.edu.cn}} \hspace{0.5cm}} \\
		{\texttt{\small{wgsg}@tsinghua.edu.cn}\hspace{0.5cm}} 
		{\texttt{\small{kugang.cwh}@alibaba-inc.com}\hspace{0.5cm}}
	}

	\maketitle
	\ificcvfinal\thispagestyle{empty}\fi
	
	\begin{abstract}
		In this work, we address the problem of unsupervised domain adaptation for person re-ID where annotations are available for the source domain but not for target. Previous methods typically follow a two-stage optimization pipeline, where the network is first pre-trained on source and then fine-tuned on target with pseudo labels created by feature clustering. Such methods sustain two main limitations. (1) The label noise may hinder the learning of discriminative features for recognizing target classes. (2) The domain gap may hinder knowledge transferring from source to target. We propose three types of technical schemes to alleviate these issues. First, we propose a cluster-wise contrastive learning algorithm (CCL) by iterative optimization of feature learning and cluster refinery to learn noise-tolerant representations in the unsupervised manner. Second, we adopt a progressive domain adaptation (PDA) strategy to gradually mitigate the domain gap between source and target data. Third, we propose Fourier augmentation (FA) for further maximizing the class separability of re-ID models by imposing extra constraints in the Fourier space. We observe that these proposed schemes are capable of facilitating the learning of discriminative feature representations. Experiments demonstrate that our method consistently achieves notable improvements over the state-of-the-art unsupervised re-ID methods on multiple benchmarks, e.g., surpassing MMT largely by 8.1\%, 9.9\%, 11.4\% and 11.1\% mAP on the Market-to-Duke, Duke-to-Market, Market-to-MSMT and Duke-to-MSMT tasks, respectively.
		
	\end{abstract}
	
	
	\begin{figure}[t]
		\centering
		\includegraphics[width=1\columnwidth]{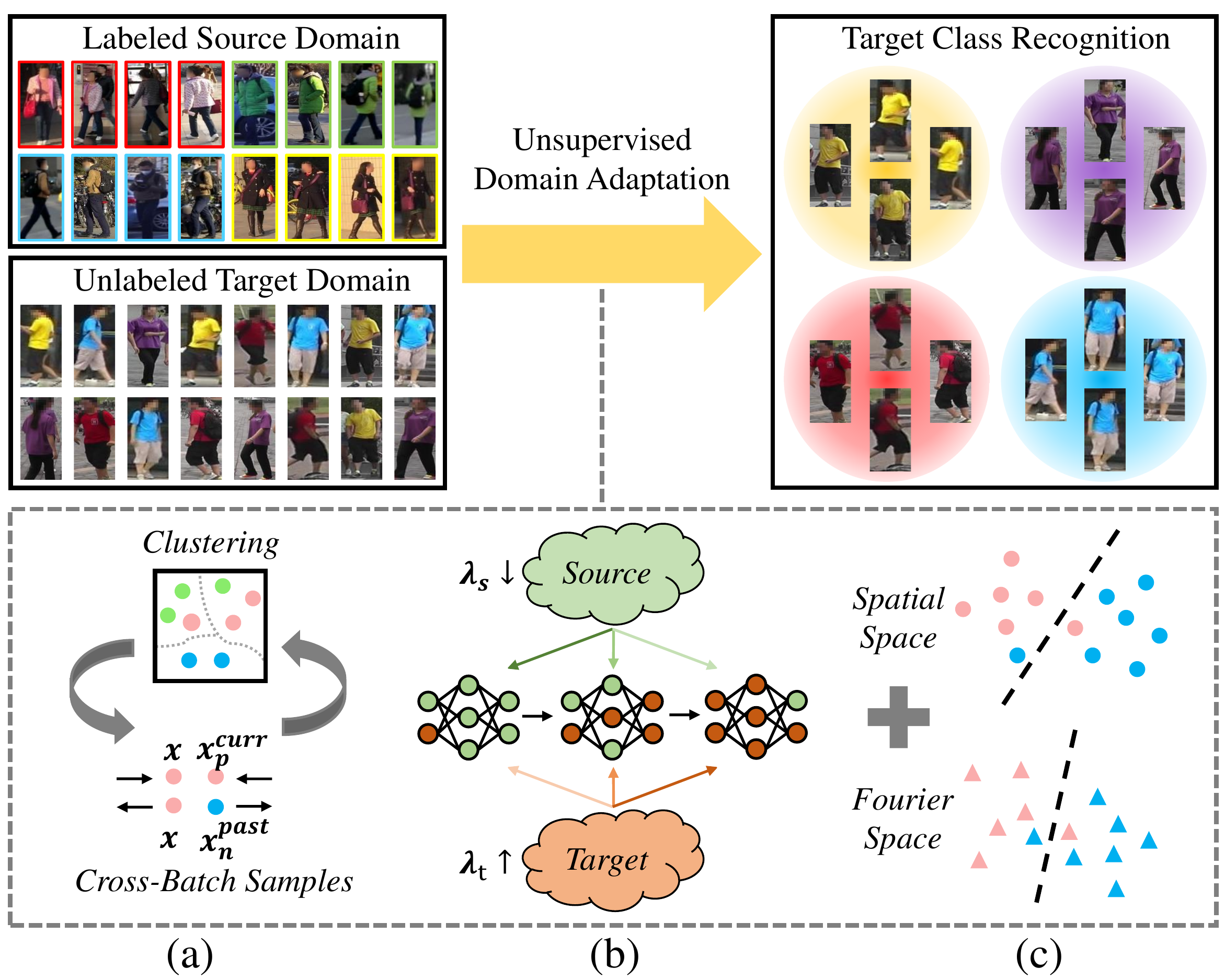}
		\vspace{2mm}
		\caption{Given labeled source data and unlabeled target data, our goal is to learn feature representations for recognizing target classes. For this unsupervised domain adaptation task, we propose three technical schemes to learn discriminative target features: (a) cluster-wise contrastive learning, (b) progressive domain adaptation (c) Fourier augmentation.}
		\vspace{-5mm}
		\label{figure:intro}
	\end{figure}
	\section{Introduction}
	
	Person re-identification (re-ID) is an important task in intelligent surveillance, which aims at identifying the person across different camera views. Recent person re-ID methods have achieved impressive performance owing to the advancement of deep convolutional neural networks (CNNs) \cite{sun2018beyond,xu2018attention,sun2017svdnet,li2018harmonious,chang2018multi,zhong2018camera,isobe2020intra}. However, the success is mainly attributed to supervised learning over massive human-labeled data. The need of time-consuming manual annotations substantially limits the scalability of re-ID models. Besides, directly applying a pre-trained re-ID model to other new domains may cause significant performance drop due to the inherent data distribution shift across different surveillance cameras. Recently, unsupervised domain adaptation (UDA) has thus attracted much attention to adapt the model learned on a labeled source domain to an unlabeled target domain.
	Prior unsupervised re-ID methods typically rely on iterative training with pseudo labels generated by clustering algorithms on the target domain \cite{zhang2019self,yang2019selfsimilarity,ge2020mutual,lin2019aBottom}. These existing methods have shown promising results but still sustain two main limitations. (1) The label noise may mislead an unexpected optimization direction for network training with the unlabeled target data. (2) The knowledge (i.e. the model ability of distinguishing person identities) learned on source can not be sufficiently transferred to target by simply fine-tuning the source model.
	
	To alleviate these problems, we investigate three aspects to facilitating the learning of discriminative feature representations for better recognizing target classes by (1) reducing the label noise on the unlabeled target data, (2) better transferring knowledge learned from source to target, and (3) adding extra training constraints. Accordingly, we propose a unified framework to achieve these goals. First, inspired by the recent contrastive learning algorithm for unsupervised visual representation learning \cite{he2019momentum,chen2020simple,wu2018unsupervised}, we propose a \emph{cluster-wise} contrastive learning algorithm to learn noise-tolerant representations on the unlabeled target data (Figure~\ref{figure:intro} (a)). Specifically, we construct a momentum-based moving-average (MMA) feature encoder and build a dynamic queue to provide sufficient negative samples across multiple mini-batches for training. Unlike the instance-wise supervision \cite{he2019momentum,chen2020simple,wu2018unsupervised}, we incorporate cluster-wise supervision generated by clustering, which is amenable for the high-level re-ID task. Our contrastive learning and feature clustering is performed in an alternating way so that the noise of pseudo labels can be largely reduced. Second, most of existing methods apply a two-stage training process in which the network is first pre-trained on source and then fine-tuned on target. Instead of directly fine-tuning, we adopt a collaborative learning mechanism on both domains with a shared feature encoder (Figure~\ref{figure:intro} (b)). By \emph{gradually} decreasing the training weights on source and increasing weights on target, we can better transfer the model ability of distinguishing person identities from source to target. Third, we propose to impose extra constraints in the Fourier space for maximizing the class separability of re-ID models (Figure~\ref{figure:intro} (c)). We view the amplitude spectrum feature as a kind of nonlinear transformations and compute additional loss functions (e.g, cross-entropy loss) to augment training. In summary, with the proposed method, we can learn better discriminative feature representations and further improve the state-of-the-art performance of unsupervised re-ID.
	
	The main contributions of this paper are summarized as follows. (1) We propose a cluster-wise contrastive learning algorithm to learn noise-tolerant feature representations on the unlabeled target data. 
	The label noise can be largely reduced in the iterative optimization procedure of feature clustering and learning. (2) We propose a progressive domain adaptation strategy to gradually transfer the knowledge learned by the labeled source domain into the unlabeled target domain for unsupervised re-ID. 
	(3) We propose to impose Fourier constraints to further maximize the class separability of the model. We observe the frequency spectrum features can be complementary to the spatial features and beneficial for improving the re-ID performance. (4) Empirical evaluations demonstrate that our method consistently outperforms prior state-of-the-art methods on multiple benchmarks by a large margin. Particularly, using the same ResNet-50 backbone, we surpass MMT \cite{ge2020mutual} by 8.1\%, 9.9\%, 11.4\% and 11.1\% mAP on the Market-to-Duke, Duke-to-Market, Market-to-MSMT and Duke-to-MSMT tasks, respectively.
	
	\section{Related Work}
	\vspace{2mm}
	
	\textbf{Unsupervised Visual Representation Learning.} Unsupervised visual representation learning aims to learn rich feature representations from large-scale unlabeled images, which is also related to our work. The key idea to perform unsupervised learning is constructing pretext tasks with free supervision. Typical methods include recovering the input image by auto-encoders \cite{vincent2008extracting,pathak2016context,zhang2016colorful}, predicting spatial context \cite{doersch2015unsupervised,noroozi2016unsupervised}, clustering features \cite{caron2019unsupervised,caron2018deep}, tracking \cite{wang2015unsupervised} or segmenting objects \cite{pathak2017learning} in videos and discriminating the instance-wise samples \cite{wu2018unsupervised,he2019momentum,chen2020simple}. Similar to \cite{he2019momentum}, we also build a contrastive self-supervised learning framework to learn representations on the unlabeled target domain. However, in contrast to relying on the instance-wise supervision by maximizing agreement between differently augmented views of the same instance, we incorporate cluster-wise supervision generated by iterative clustering into contrastive learning. We observe that the class information is more suitable to learn discriminative representations for the re-ID task. Moreover, rather than collecting all the samples from the queue \cite{he2019momentum}, we filter out those having the same pseudo-class with the anchor to ensure the quality of negative samples. Contrastive learning is also widely used in many supervised learning methods where training samples are off-the-shelf with labels. In this work, we focus on how to collect meaningful pairs and reduce the label noise in the unsupervised case.
	
	\textbf{Unsupervised Domain Adaptation.} Generic unsupervised domain adaptation (UDA) methods address the closed-set problem where the target domain shares the same semantic classes with the source domain. Typical UDA methods focus on reducing the domain discrepancy by aligning data distribution between source and target domains \cite{yan2017mind,shen2018wasserstein,zellinger2017central,saito2018maximum,sohn2018unsupervised}, training adversarial domain-classifiers to encourage features of source and target domains to be indistinguishable \cite{zhang2018collaborative,tzeng2017adversarial,isobe2021multi}, or learning domain-specific properties \cite{bousmalis2016domain,gong2013connecting,long2015learning,zhang2018collaborative,chang2019domain}. In this work, we tackle the more challenging open-set problem of UDA for re-ID, where the classes between the source and target domains are not shared. In fact, our method does not rely on any assumption on the classes. The classes between source and target can be exactly the same, totally different or partially shared.
	
	\textbf{Unsupervised Cross-Domain Person Re-ID.} Although supervised person re-ID methods have achieved great performance on the trained data domain \cite{he2018deep,sun2018beyond,xu2018attention,paisitkriangkrai2015learning,sun2017svdnet,li2018harmonious,chang2018multi,zhong2018camera,he2019foreground,he2021transreid,zhang2021graph}, the accuracy often drops significantly when directly testing on a different domain. Recently, unsupervised cross-domain person re-ID methods~\cite{chang2018disjoint,deng2018image,song2018unsupervised,lin2018multi,wang2020camera,zheng2021group} have attracted much attention to address the problem. Typical approaches~\cite{zhang2019self,yang2019selfsimilarity,qi2019novel,ge2020mutual,yang2021joint} take a pre-trained model on the labeled source domain as the initialized feature encoder, and further optimize it on the unlabeled target domain by metric learning or unsupervised clustering. Instead of directly fine-tuning the source model, we progressively transfer the knowledge from source to target. 
	Some approaches \cite{ge2020mutual,zhong2019invariance,yu2019unsupervised} apply soft labels for training on target, which can reduce the effect of noise to some extent in the optimization process. NRMT~\cite{zhao2020unsupervised} introduces a
	collaborative clustering to fit to noisy instances.
	Another line of recent work~\cite{zhai2020ad} attempts to learn domain-invariant features from style-transferred images. DG-Net++~{zou2020joint} disentangles
	feature space from each domain into id-related
	and id-unrelated components However, the model performance heavily counts on the image generation quality and how to optimize the class separability of learned representations is often neglected. Recent work of \cite{ge2020self,zheng2020exploiting} jointly optimizes both source and target domains to produce reliable pseudo labels. Our work is related to~\cite{ge2020self} in the aspect of contrastive learning. The main differences are three-fold. (1) For the contrastive loss, \cite{ge2020self} integrates instance-level, cluster-level and class-level supervision on both domains while we only employ cluster-level supervision on the target data. Besides, we do not rely on additional tricks to select clusters (e.g., independence or compactness used in \cite{ge2020self}). (2) \cite{ge2020self} also performs joint learning of source and target domains. Differently, we focus on progressive training by gradually decreasing the source weights and increasing the target weights. (3) \cite{ge2020self} only relies on conventional spatial features while we propose to add Fourier constraints for improving the class separability of re-ID models.
	
	\vspace{2mm}
	\textbf{Learning in Fourier Space.} The discrete Fourier transform (DFT) converts a finite sequence of values into components of different frequencies, which is a classical mathematical transform method and has many practical applications such as digital signal processing and image processing. Its fast computation algorithm of fast Fourier transform (FFT) and the variant of discrete cosine transform (DCT) have been widely used in data compression. Recently, Fourier transform has also been explored for compressing CNNs by grouping the frequency coefficients of kernel weights into hash buckets \cite{chen2016compressing}, discarding the low-energy frequency coefficients \cite{wang2018packing} or training with band-limited frequency spectra \cite{dziedzic2019band}. Another line of recent work attempts to learn straight from the compressed representations in the Fourier space \cite{torfason2018towards,gueguen2018faster,ehrlich2019deep,xu2020learning} for efficient training and inference. To reduce the discrepancy between the source and target distributions, FDA \cite{yang2020fda} adopts style transfer in Fourier space by swapping the low-frequency spectrum of source with the target, which shows promising results on semantic segmentation. However, image-level perceptual changes can cause significant deterioration of the performance of person re-ID since it relies heavily on the appearance characteristics of persons. Different from prior methods, we apply 1D FFT for converting the output of network from spatial space to Fourier space, and then combine the losses that computed on the FFT features for network optimization. 
	
	\begin{figure}[t]
		\centering
		\includegraphics[width=1\columnwidth]{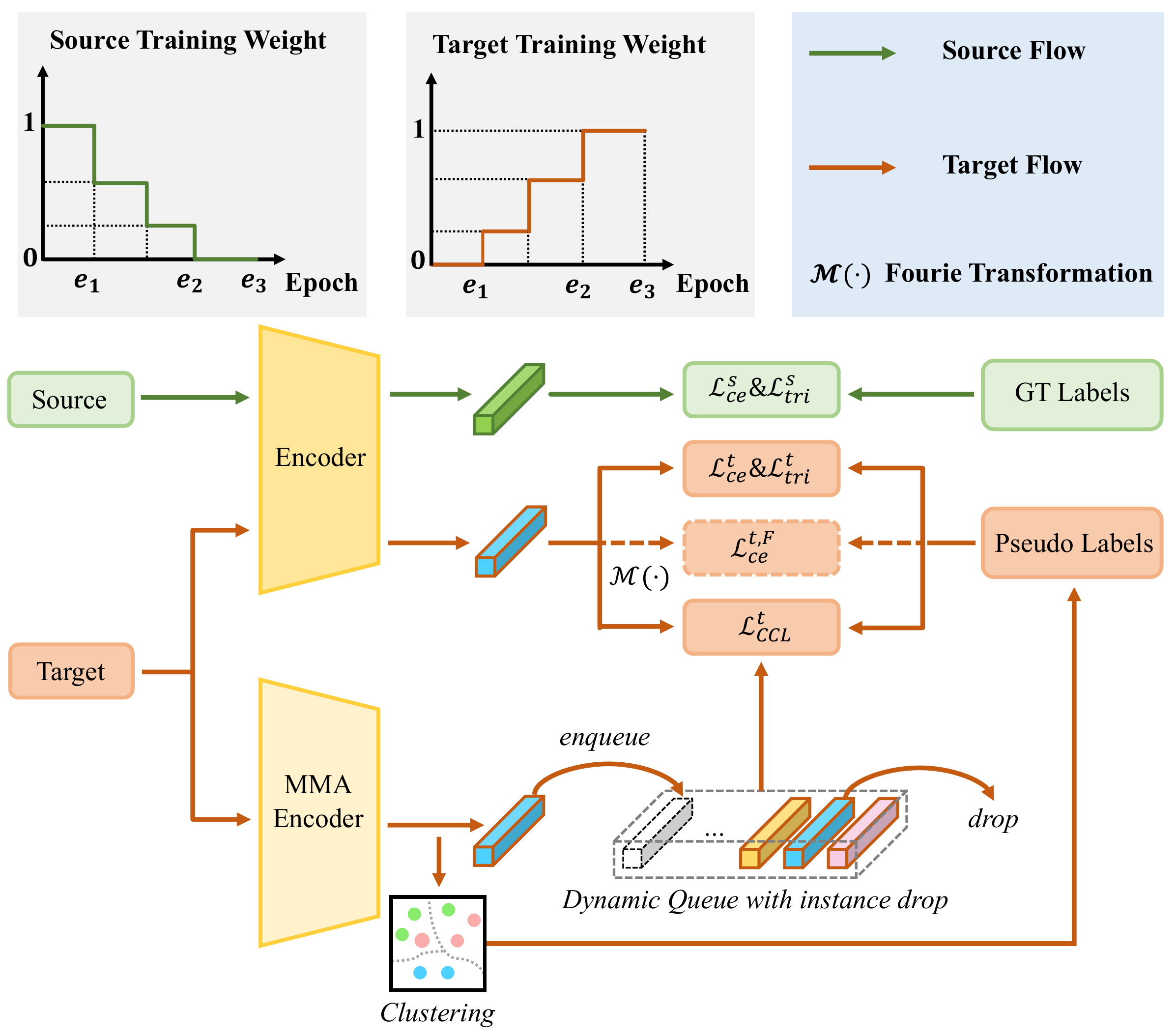}
		\vspace{3mm}
		\caption{Illustration of the proposed unified framework for unsupervised cross-domain person re-ID.}
		\vspace{-5mm}
		\label{framework}
		\label{figure:overview}
	\end{figure}
	
	\section{Methodology}
	\label{sec:method}
	We denote the source domain data as  $\mathcal{D}_s=\{(x^s_i,y^s_i)\}^{N_s}_{i=1}$, where $x^s_i$ and $y^s_i$ indicate the $i$-th training samples and its corresponding class label. The target domain is denoted as $\mathcal{D}_t=\{x^t_i\}^{N_t}_{i=1}$, where class labels are not available. The goal of unsupervised cross-domain person re-ID is to learn a mapping function $f_\theta(\cdot)$ to identify the class label (i.e., person identity) for each target image, where $\theta$ is the parameters to be learned. The general optimization target can be formulated as: 
	\begin{equation}
		\mathcal{L}(\theta)=\lambda_\text{s}(e) \cdot \mathcal{L}^s(\theta) + \lambda_\text{t}(e) \cdot \mathcal{L}^t(\theta),
		\label{source-target}
	\end{equation}
	where $\mathcal{L}^s$ and $\mathcal{L}^t$ indicate the optimization targets on the source and target domains, respectively. $\lambda_s(e)$ and $\lambda_t(e)$ are variables that change over time ($e$ means epoch) to control training on the source and target domains, respectively. 
	
	Most of existing methods adopt a two-stage optimization pipeline to address this task. That is, the model is first pre-trained on source using the ground-truth labels and then fine-tuned on target using pseudo labels generated by clustering. We formulate the two-stage baseline method as:
	\begin{equation}
		\begin{aligned}
			\mathcal{L}(\theta) = {} & \lambda_\text{s}(e) \cdot (\mathcal{L}_\text{ce}(\theta; y^s) + \mathcal{L}_\text{tri}(\theta; y^s)) \\ & + \lambda_\text{t}(e) \cdot (\mathcal{L}_\text{ce}(\theta; \hat{y}^t) + \mathcal{L}_\text{tri}(\theta; \hat{y}^t)) ,
			\label{baseline}
		\end{aligned}
	\end{equation}
	\begin{small}
		\begin{equation}
			\lambda_\text{s}(e) = \left\{\begin{matrix} 1 & e \in (0,e_1] \vspace{2mm} \\ 0 & e \in (e_1,e_2]
			\end{matrix}\right., 
			\
			\lambda_\text{t}(e) = \left\{\begin{matrix} 0 & e \in (0,e_1] \vspace{2mm} \\ 1 & e \in (e_1,e_2]
			\end{matrix}\right.
			\label{2-stage}
		\end{equation}
	\end{small}
	
	where $\mathcal{L}_{ce}$ and $\mathcal{L}_{tri}$ represents the cross-entropy classification and triplet loss~\cite{hermans2017defense}, respectively. $y^s$ and $\hat{y}^t$ means the ground-truth class labels on source and pseudo labels on target, respectively. 
	
	However, the pseudo labels generated by clustering inevitably contain noise (i.e., wrong labels), which may cause wrong optimization directions during network training.
	
	\begin{figure}[t]
		\centering
		\includegraphics[width=1\columnwidth]{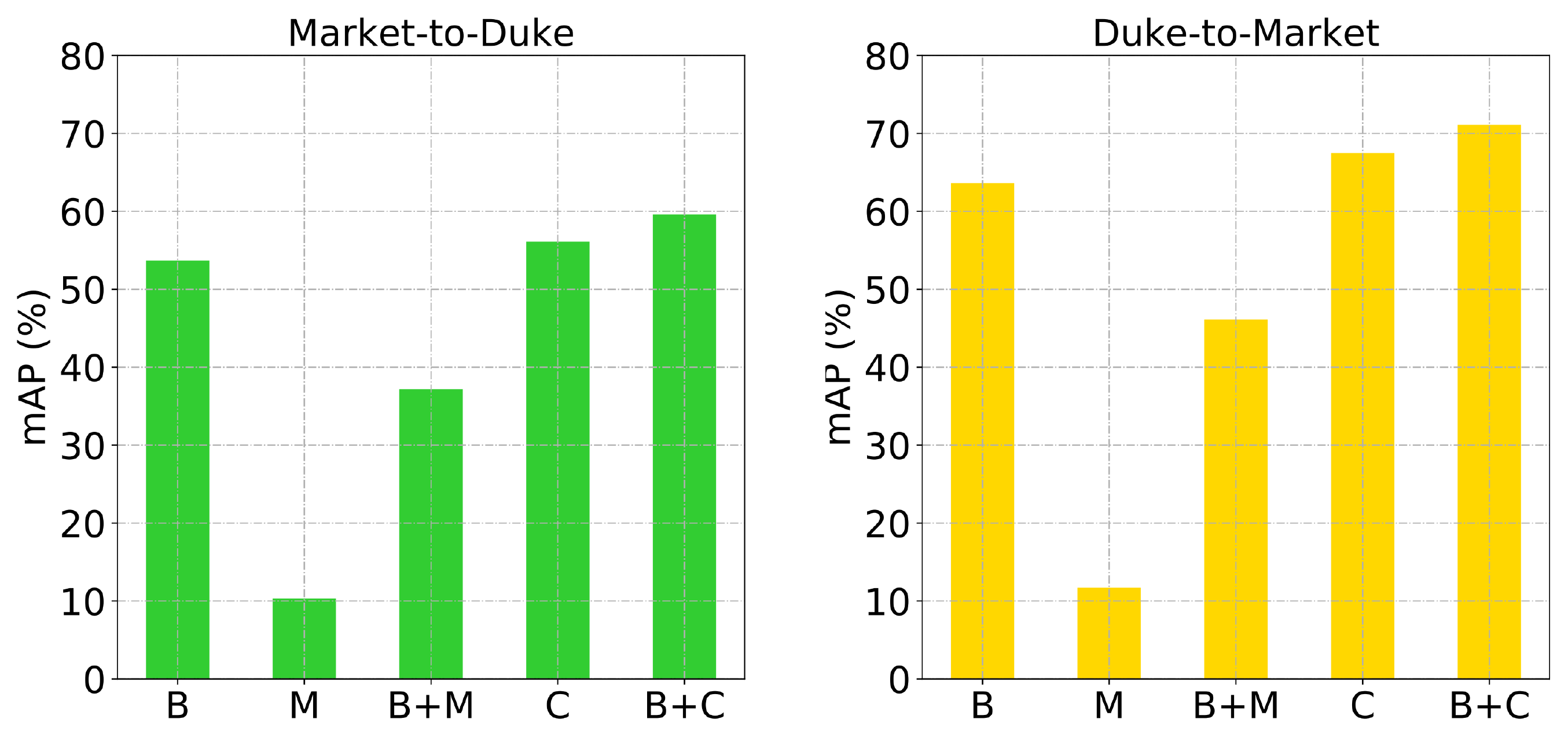}
		\vspace{2mm}
		\caption{For a fair comparison, we first pre-train the networks on the labeled source data, and then fine-tune these networks with different contrastive learning methods on target. "B", "M" and "C" represents "Baseline", "MoCo" and "CCL", respectively.}
		\vspace{-8mm}
		\label{figure:ccl}
	\end{figure}
	
	\subsection{Cluster-wise Contrastive Learning}
	In order to reduce label noise on the unlabeled target data, we propose a cluster-wise contrastive learning algorithm, which is inspired by the recent success of unsupervised feature learning \cite{he2019momentum,wang2020cross}. In detail, we perform feature clustering and contrastive learning in an alternating manner, towards refining the noisy pseudo labels and updating network weights iteratively. In each round of alternating training, we first employ unsupervised feature clustering (e.g., DBSCAN) to generate pseudo labels and design a cluster-wise contrastive loss to train the network:
	\begin{equation}
		\mathcal{L}^t_\text{ccl}(\theta) = - \log \frac{\exp{(f_\theta(x^t)} f_\theta(x_p^t) / \tau)}{\sum_{x_n \in \mathbb{N}_\text{past}}\exp{(f_\theta(x^t)} f_{\hat\theta}(x_n^t) / \tau)},
		\label{eq:ccl}
	\end{equation}
	where $x_p^t$ and $x_n^t$ indicate the positive (i.e., same pseudo class) and negative (i.e., different pseudo classes) samples to $x^t$, respectively. $\hat\theta$ represents a momentum-based moving-averaging (MMA) feature encoder to maintain consistency between past and current features, which is updated as $\hat\theta \gets m\hat\theta + (1-m)\theta$ during training. Here, $m$ is a momentum coefficient which is set to 0.99 in our method. We only update the parameters of the regular encoder $\theta$ by back-propagation and constrain no gradient back-propagating for the MMA encoder $\hat\theta$. For the anchor $x^t$, we select its positive sample within the current batch and collect past negative samples from previous batches. To implement cross-batch sampling, we build a dynamic queue to memorize the past features $\mathbb{N}_\text{past}$. However, it is unreasonable to simply take all the past features as negative samples because previous batches may contain positive samples of the anchor. Hence, based on pseudo labels generated in the current round, we drop the instances which have the same class with the anchor from the queue to ensure the quality of negative samples. By collecting sufficient negative samples across multiple batches for training, the network can be better optimized compared to a naive contrastive loss with very limited training samples within a single batch.

	\begin{figure}[t]
		\centering
		\setlength{\belowcaptionskip}{-0.2cm}
		\subfigure[Person re-ID Performance]{
			\label{fig21}
			\includegraphics[width=1.55in]{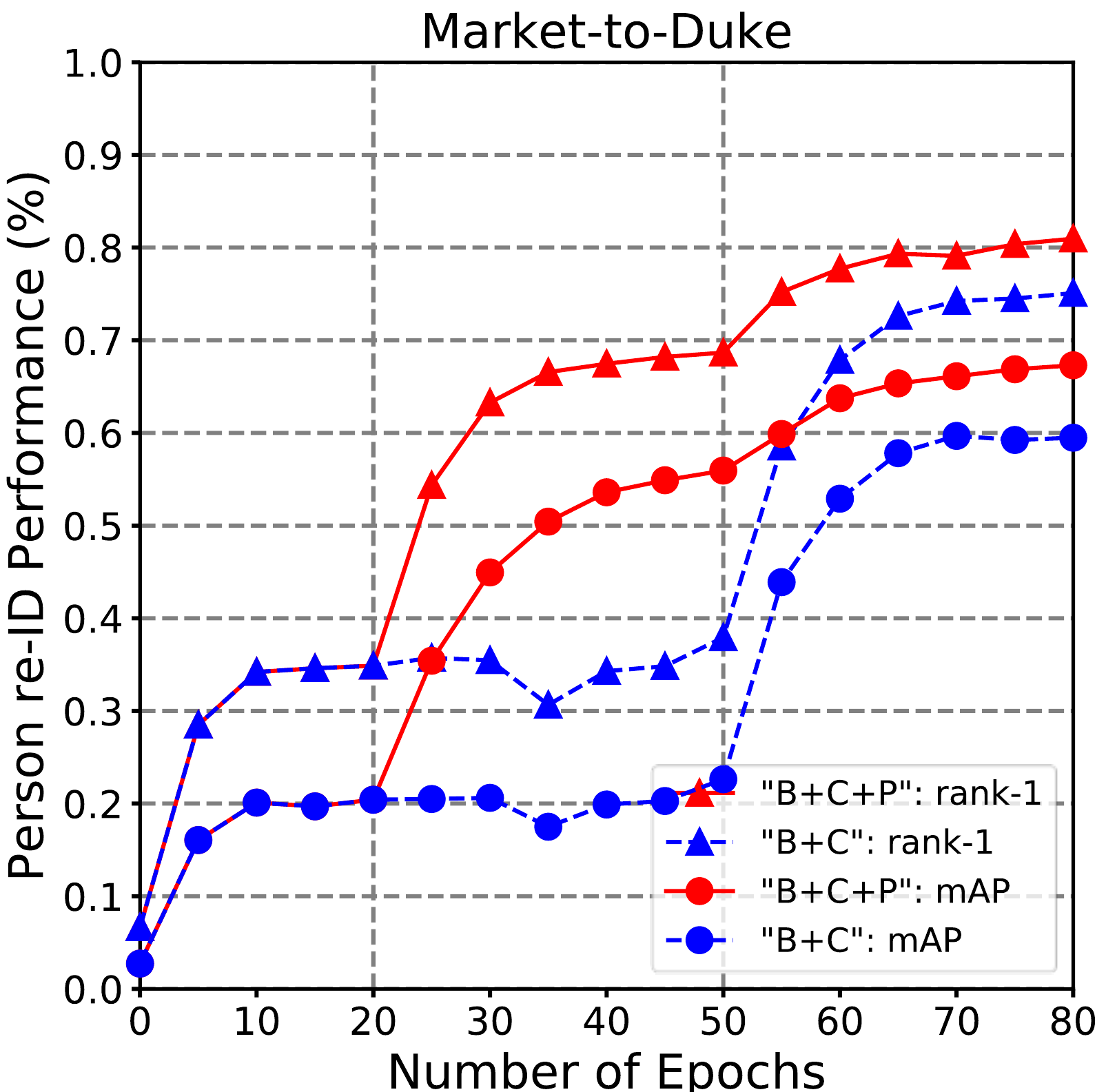}}
		\subfigure[Clustering Performance]{
			\label{fig22}
			\includegraphics[width=1.55in]{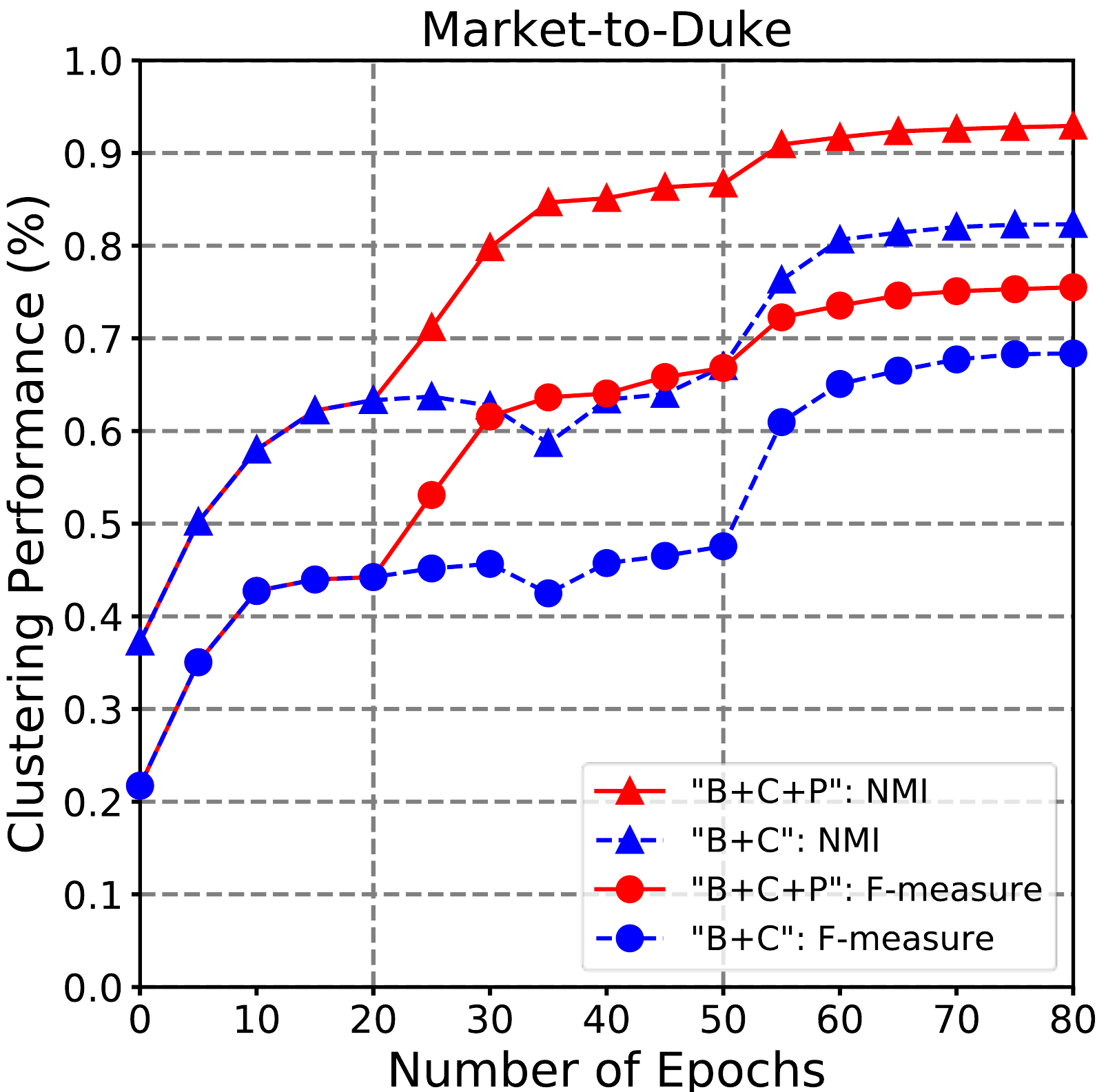}}
		\vspace{2mm}
		\caption{Illustration of the proposed progressive domain adaptation (PDA) in term of the re-ID and clustering performance. (a) the re-ID performance (mAP and rank-1 accuracy) on the test set. (b) the clustering performance (NMI \cite{wang2019linkage} and F-measure \cite{  amigo2009comparison}) on the training set.}
		\vspace{-5mm}
		\label{figure:curve}
	\end{figure}

	We summarize the main differences between CCL with MoCo \cite{he2019momentum} as follows. First, MoCo relies on the instance-wise supervision by maximizing agreement between differently augmented views of the same instance, while we extend it to a cluster-wise version by exploiting the pseudo labels generated by clustering to construct pairs for learning. We observe that such class information is more suitable for the re-ID task. Second, in contrast to keep updating the queue throughout the training process, we refresh the queue when pseudo labels are updated by a new round of feature clustering, owing to the fact of the class label of a specific training sample is not consistent across different clustering procedures. 
	
	Figure~\ref{figure:ccl} reinforces our intuition that cluster-wise supervision is crucial for the re-ID performance. MoCo almost fails on both benchmarks. By combining with the baseline (Eq. \ref{baseline}), MoCo still yields inferior results. One can reason that instance-level supervision and its optimization target of MoCo is different from the re-ID task. Training with instance-wise pairs may hinder the learning of feature representations to distinguish different classes. We also find that CCL can boost the baseline by a large margin on both benchmarks, validating the non-trivial design and effectiveness of our method.
	
	\subsection{Progressive Domain Adaptation}
	Individually training each domain (Eq.~\ref{2-stage}) is not optimal for knowledge transfer, especially when there is a large discrepancy between source and target domains. Moreover, when the number of labeled source images is limited, it can easily induce an overfitting trap and hamper the knowledge transfer from source to target.
	
	To alleviate this problem, we propose a progressive domain adaptation strategy to gradually optimize $\mathcal{L}(\theta)$ from source to target. Specifically, we decrease the source training weights and increase the target training weights over time. Unlike the two-stage training baseline (Eq.~\ref{2-stage}), we can formulate $\lambda_\text{s}$ and $\lambda_\text{t}$ as:
	\begin{scriptsize}
		\begin{equation}
			\lambda_\text{s}(e) = \left\{\begin{matrix} 1,\quad e \in (0,e_1] \vspace{2mm}  \\ w(e), e \in (e_1,e_2] \vspace{2mm} \\ 0,\quad e \in (e_2,e_3] 
			\end{matrix}\right.,  
			\
			\lambda_\text{t}(e) = \left\{\begin{matrix} 0,\qquad e \in (0,e_1] \vspace{2mm} \\ 1-w(e), e \in (e_1,e_2] \vspace{2mm} \\ 1,\qquad e \in (e_2,e_3]
			\end{matrix}\right.
			\label{eq:pda}
		\end{equation}
	\end{scriptsize}
	where $w(e)$ defines a decay policy. For example, a multi-step policy can be illustrated in Figure \ref{figure:overview}. The training process is divided into three phases according to Eq. \ref{eq:pda}. First, we follow common practice in supervised re-ID to pre-train the model on source  ($\mathcal{L}(\theta)=\mathcal{L}^s(\theta)$) as an initialization for the subsequent optimization. Second, we jointly train the network on both source and target domains. For the labeled source data, the optimization objective remains the same as the pre-training phase. For the unlabeled target data, the optimization objective is combination of CCL, cross-entropy and triplet losses based on pseudo labels. Third, since our goal is to accurately predict the target classes as possible, we continue to train the network with the target data only in the final phase ($\mathcal{L}(\theta)=\mathcal{L}^t(\theta)$). Figure \ref{figure:curve} shows the clustering and re-ID performance throughout training. According to the clustering performance, the results imply that our method can gradually reduce label noise and yield cleaner clusters compared to the two-stage baseline. According to the re-ID performance, the results imply that our method can learn better features gradually and achieve higher recognition performance than the baseline.

	\subsection{Fourier Augmentation}
\begin{figure}[t]
		\centering
		\includegraphics[width=1\columnwidth]{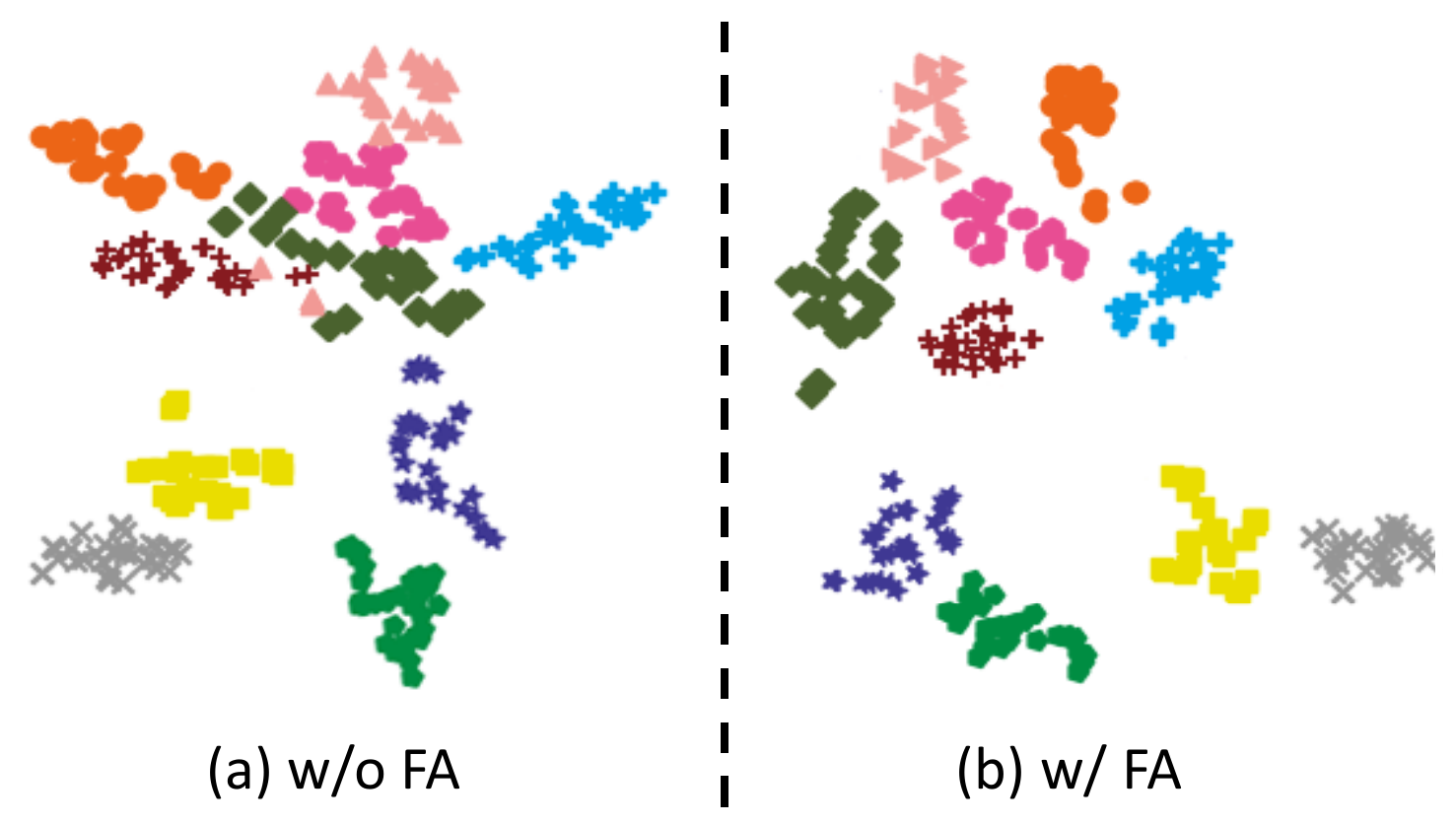}
		\vspace{-3mm}
		\caption{Analysis of the proposed Fourier augmentation (FA) in feature space using t-SNE~\cite{van2008visualizing} visualization. \textbf{(Zoom-in for best view)}}
		\vspace{-5mm}
		\label{FFT-analysis}
	\end{figure}

	Inspired by~\cite{tancik2020fourier}, we consider to impose extra optimization constraints in Fourier space. Specifically, we first apply Fast Fourier Transform (FFT) to compute the real and imaginary components of the 1D CNN output features. We then exploit the amplitude spectrum $\mathcal{M}(x) = ||\mathcal{F}(f_\theta(x))||$ to compute the cross-entropy loss for the target data. In order to better understand the proposed Fourier augmentation scheme, we provide analysis in the following aspects. (1) $\mathcal{M}(\cdot)$ can be viewed as a kind of nonlinear mappings. Thus, joint training on the spatial and Fourier features implies that loss functions are computed for different nonlinear features of a training image. We empirically find that such amplitude spectrum features performs better than an extra single MLP layer. (2) Figure \ref{FFT-analysis} visualizes the CNN feature distribution with and without our Fourier augmentation. he qualitative results show that by imposing these extra constrains for training, different classes can be better distinguished. (3) In mathematics, Parseval's Theorem (a special case of Plancherel Theorem) states the relation between a signal and its Fourier transform. In our case, the relation becomes:
	\begin{equation}
		\| f_\theta(x) \|^2 = \frac{1}{D} \| \mathcal{M}(f_\theta(x)) \|^2 
	\end{equation}
	where $D$ means the feature length. According to this nature, the triplet losses based on Euclidean distance will be equivalent for the spatial and Fourier features. Hence, we only add the cross entropy loss in the Fourier space.

	\subsection{Overall Optimization}
	The overall optimization objective of our method can be defined as:
	\begin{equation}
		\begin{footnotesize}
			\vspace{2mm} \\
			\mathcal{L}(\theta) = \lambda_\text{s}(e) \cdot \mathcal{L}^s + \lambda_\text{t}(e) \cdot (\delta \mathcal{L}_\text{ccl}^t + \gamma \mathcal{L}_\text{spa}^t + (1-\gamma) \mathcal{L}_\text{fre}^t),
		\end{footnotesize}
	\end{equation}
	where $\gamma$ is a loss weight that balances the spatial and Fourier losses. $\delta$ controls the effect of cluster-wise contrastive learning. We compute both of the cross-entropy and triplet losses for $\mathcal{L}^s$ and $\mathcal{L}_\text{spa}^t$ and only compute the cross-entropy loss for $\mathcal{L}_\text{fre}^t$.
	
	\section{Experiments}
	\label{experiment}
	\begin{table*}[t]
		\scriptsize
		\centering
		\begin{center}
			\scalebox{1}{
				\begin{tabular}{P{3.0cm}|C{1.2cm}|C{0.7cm}C{0.7cm}|C{0.7cm}C{0.7cm}|C{0.7cm}C{0.7cm}|C{0.7cm}C{0.7cm}} 
					\toprule
					\multicolumn{2}{c|}{\multirow{2}{*}{Method}} & \multicolumn{2}{c|}{Market-to-Duke} & \multicolumn{2}{c|}{Duke-to-Market} & \multicolumn{2}{c|}{Market-to-MSMT} & \multicolumn{2}{c}{Duke-to-MSMT}\\
					\cmidrule{3-10}
					\multicolumn{2}{c|}{} & mAP & rank-1 & mAP & rank-1 & mAP & rank-1   & mAP & rank-1  \\ 
					\midrule
					PUL~\cite{fan18unsupervisedreid} &TOMM'18 & 16.4 & 30.0 & 20.5 & 45.5 &- &- &- &-  \\
					HHL~\cite{zhong2018generalizing} &ECCV'18 & 27.2 & 46.9  & 31.4 & 62.2 &- &- &- &-   \\
					PTGAN~\cite{wei2018person} &CVPR'18 &-  &27.4 &- &38.6  &2.9 &10.2  &3.3 &11.8 \\
					TJ-AIDL~\cite{wang2018transferable} &CVPR'18 & 23.0 & 44.3  & 26.5 & 58.2 &- &- &- &-  \\
					ARN~\cite{li2018adaptation} &CVPRW'18 & 33.4 & 60.2  & 39.4 & 70.3 &- &- &- &-  \\
					MMFA~\cite{lin2018multi} &BMVC'19 &24.7 &45.3  &27.4 &56.7  & &  & & \\
					PDA-Net~\cite{li2019cross} &ICCV'19 & 45.1 & 63.2  & 47.6 & 75.2  &- &- &- &-  \\
					PCB-PAST~\cite{zhang2019self} &ICCV'19 & 54.3 & 72.4  & 54.6 & 78.4 &- &- &- &- \\
					SSG~\cite{yang2019selfsimilarity} &ICCV'19 & 53.4 & 73.0  & 58.3 & 80.0  &13.2 &31.6  &13.3 &32.2 \\
					CR-GAN~\cite{chen2019instance}  &ICCV'19 &48.6 &84.7   &54.0 &77.7  &- &- &- &-   \\
					ECN++~\cite{zhong2020memory}  &TPAMI'20 &54.4 &74.0  &63.8 &84.1  &15.2 &40.4  &16.0 &42.5  \\
					MMCL~\cite{wang2020unsupervised} &CVPR'20 &51.4 &72.4    &60.4 &84.4 &15.1 &40.8  &16.2 &43.6  \\
					SNR~\cite{jin2020style} &CVPR'20 &58.1 &76.3   &61.7 &82.8 &- &- &- &-  \\
					DG-Net++~\cite{zou2020joint} &ECCV'20 &63.8 &78.9  &61.7 &82.1 &- &- &- &-  \\
					NRMT~\cite{zhao2020unsupervised} &ECCV'20 &62.2 &77.8 &71.7  &87.8 &19.8 &43.7  &20.6 &45.2 \\
					MEB-Net~\cite{zhai2020multiple}&ECCV'20 &66.1 &79.6 &76.0 &89.9  &- &- &- &- \\
					MMT ($k$-means)~\cite{ge2020mutual} &ICLR'20 &65.1  &78.0  & 71.2 &87.7  &22.9 &49.2  &23.5 &50.1  \\
					MMT(DBSCAN)$^\dagger$~\cite{ge2020mutual} &ICLR'20 &62.7  &76.8  &73.5 &89.7 &24.4 &50.7  &25.2 &53.2 \\
					SpCL~\cite{ge2020self} &NeurIPS'20 &-  &-  &-  &-  &26.8 &53.7  &- &- \\
					\rowcolor[gray]{.92}Baseline &Ours  &53.7 &69.9 &63.6  &82.5 &14.5 &33.3 &17.1 &38.4   \\
					\rowcolor[gray]{.92}Baseline + CCL  & Ours  &59.6  &75.0 &71.1 &87.8 &20.1 &42.7 &22.9 &48.4 \\
					\rowcolor[gray]{.92} Baseline + CCL + PDA  &Ours   &67.3  &80.9 &80.3 &92.5 &30.7 &59.0 &30.1 &59.5  \\
					\rowcolor[gray]{.92} Baseline + CCL + PDA + FA  &Ours  &\textbf{69.4}  &\textbf{82.7}  &\textbf{82.2} &\textbf{93.6}  &\textbf{32.9} &\textbf{61.8} &\textbf{32.7} &\textbf{62.7} \\
					\rowcolor[gray]{.92} Baseline + CCL + PDA + FA  &~~Ours* &\textbf{70.8}  &\textbf{83.5}  &\textbf{83.4} &\textbf{94.2}  &\textbf{35.8} &\textbf{65.8} &\textbf{36.3} &\textbf{66.6}  \\
					\midrule
					\multicolumn{2}{c|}{Supervised baseline } &72.3 &84.4 &82.8 &93.6 &44.7 &72.4  &44.7  &72.4 \\
					\multicolumn{2}{c|}{Supervised baseline + FA } &74.4 &86.0  &84.5 &94.8 &47.1 &75.2  &47.1 &75.2 \\
					\bottomrule
				\end{tabular}
			}
		\end{center}
		\vspace{2mm}
		\caption{Performance comparisons on multiple benchmarks for unsupervised cross-domain person re-ID. The supervised baseline is obtained by training with cross-entropy and triplet losses using the ground-truth labels of target data. More stronger supervised baseline is obtained by combining the loss in Fourier space. $^\dagger$ means our re-implementation of \cite{ge2020mutual} with the DBSCAN clustering algorithm for fair comparisons. The results of "Ours*" are obtained by combining the proposed method with soft cross-entropy loss, soft triplet loss and mutual learning strategy introduced by \cite{ge2020mutual}.
		}
		\label{tab:sota}
		\vspace{-5mm}
	\end{table*}
	
	\begin{table}[t]
		\scriptsize
		\centering
		\begin{center}
			\scalebox{1}{
				\begin{tabular}{P{1.3cm}|C{0.7cm}C{0.7cm}|C{0.7cm}C{0.7cm}}
					\toprule
					\multicolumn{1}{c|}{\multirow{2}{*}{Method}} & \multicolumn{2}{c|}{PersonX-to-Market} & \multicolumn{2}{c}{PersonX-to-MSMT} \\
					\cmidrule{2-5}
					& mAP & rank-1 & mAP & rank-1 \\ 
					\midrule
					MMT~\cite{ge2020mutual}  &70.7 &86.2 &18.2 &39.5 \\
					SpCL~\cite{ge2020self}  &73.8 &88.0 &22.7 &47.7\\
					\midrule
					Ours &\textbf{78.4}   &\textbf{91.3} &\textbf{26.2} &\textbf{50.1}\\
					Ours*  &\textbf{79.6} &\textbf{92.5} &\textbf{28.9} &\textbf{53.2}\\
					\bottomrule
				\end{tabular}
			}
		\end{center}
		\vspace{2mm}
		\caption{Comparison with the state-of-the-art unsupervised re-ID methods under the synthetic-to-real setting. }
		\label{tab:synthetic}
		\vspace{-5mm}
	\end{table}
	
	\begin{table}[t]
		\scriptsize
		\centering
		\begin{center}
			\scalebox{1}{
				\begin{tabular}{P{1.3cm}|C{0.7cm}C{0.7cm}|C{0.7cm}C{0.7cm}}
					\toprule
					\multicolumn{1}{c|}{\multirow{2}{*}{Method}} & \multicolumn{2}{c|}{VehicleID-to-VeRi} & \multicolumn{2}{c}{VehicleX-to-VeRi} \\
					\cmidrule{2-5}
					& mAP & rank-1 & mAP & rank-1 \\ 
					\midrule
					MMT~\cite{ge2020mutual} &35.3 &74.6 &35.6 &76.0 \\
					SpCL~\cite{ge2020self}  &38.4 &79.9 &38.3 &82.1 \\
					\midrule
					Ours &\textbf{41.2}   &\textbf{83.6} &\textbf{41.4} &\textbf{85.3}\\
					Ours*  &\textbf{42.7} &\textbf{84.7} &\textbf{42.5} &\textbf{86.5}\\
					\bottomrule
				\end{tabular}
			}
		\end{center}
		\vspace{-3mm}
		\caption{Comparison with other unsupervised domain adaptation methods for Vehicle re-ID tasks. The results of MMT are taken from~\cite{ge2020self}.}
		\label{tab:vehicle}
		\vspace{-3mm}
	\end{table}
	
	\begin{table}[t]
		\scriptsize
		\centering
		\begin{center}
			\scalebox{1.2}{
				\begin{tabular}{P{1.3cm}|C{0.7cm}C{0.7cm}|C{0.7cm}C{0.7cm}}
					\toprule
					\multicolumn{1}{c|}{\multirow{2}{*}{Method}} & \multicolumn{2}{c|}{Market-to-Duke} & \multicolumn{2}{c}{Duke-to-Market} \\
					\cmidrule{2-5}
					& mAP & rank-1 & mAP & rank-1 \\ 
					\midrule 
					SupCon$^\dagger$~\cite{khosla2020supervised} &66.0 &79  .4 &75.4 &88.1 \\
					\midrule
					InstDisc$^\dagger$~\cite{wu2018unsupervised} &1.9 &4.1 &2.4 &5.9\\
					MoCo$^\dagger$~\cite{he2019momentum}  &10.3 &17.7 &11.7 &26.2\\ \midrule
					CCL (Ours)  &\textbf{56.8} &\textbf{71.9} &\textbf{67.5} &\textbf{84.2}\\
					\bottomrule
				\end{tabular}
			}
		\end{center}
		\vspace{2mm}
		\caption{Performance comparisons with other contrastive learning methods and our method. "$\dagger$" means our implementation based on the official code. The cross-entropy and triplet losses are not used for all the experiments here.}
		\label{tab:ablation-CL}
		\vspace{-5mm}
	\end{table}
	
	\subsection{Datasets}
	We evaluate the proposed PDA algorithm on three real-world person re-ID datasets: DukeMTMC-reID~\cite{dukemtmc}, Market-1501~\cite{market} and MSMT17~\cite{wei2018person}. 
	The DukeMTMC-reID dataset contains 1,812 identities with 36,411 images captured by eight cameras, which splits 702 identities with 16,522 images for training and the remaining images for testing.
	The Market-1501 dataset consists of 32,688 images of 1,501 identities captured by six cameras, where the training set contains 12,936 images of 751 identities, and the test set contains 19,732 images of 750 identities. The MSMT17 dataset is a large-scale person re-ID dataset, which consists of 126,441 bounding boxes of 4,101 identities captured by fifteen cameras, for which 32,621 images of 1,041 identities are used for training. We report performance on four real-world unsupervised domain adaptation tasks: Duke-to-Market, Market-to-Duke, Duke-to-MSMT and Market-to-MSMT, where the ground-truth labels are provided on source only. We also conduct domain adaptation experiments under the synthetic-to-real setting, where PersonX~\cite{sun2019dissecting} is used as the synthetic source domain. We use the standard mean average precision (mAP) and cumulative matching characteristics (CMC) at rank-1 accuracy as evaluation metrics.

	\subsection{Implementation Details}
	We use the network (e.g., ResNet-50~\cite{he2016deep}) pre-trained on ImageNet~\cite{deng2009imagenet} as the initial feature encoder. We first train the network for $e_1=20$ epochs on source. For progressive training on both source and target domains, we adopt a $k$-step policy where the loss weights of source and target are decreased and increased by $k$ times, respectively. We train $30$ epochs (i.e., $e_2=50$) in this phase. To learn discriminative features as possible to distinguish target classes, we continue to optimize the model with the target data only in another $30$ epochs (i.e., $e_3=80$). The output of MMA encoder is used for inference. For hyper-parameters $\delta$ and $\gamma$, we conduct parameter analysis to obtain the best choices ($\delta=0.1$, $\gamma=0.7$) on the Market-to-Duke benchmark and fix them on the other benchmarks. For cluster-wise contrastive learning, we set the temperature parameter $\tau$ as $0.07$~\cite{he2019momentum} and set the maximum length of the queue as 1024. The training data are cropped to $256\times 128$ and augmented by flipping and rotating with a probability of 0.5. The network is optimized by Adam optimizer with $\beta_1 = 0.9$, $\beta_2=0.999$ and weight decay of $5\times10^{-4}$. We set a constant learning rate of $3.5\times10^{-3}$ for the entire training process. All of our experiments are conducted on PyTorch 1.1 with 4 TITAN XP GPUs. It costs 8, 10, 15, 15 hours for training our models on Duke-to-Market, Market-to-Duke, Market-to-MSMT, Duke-to-MSMT, respectively. 
	
	
	\subsection{Comparisons to the State-of-the-Arts}
	
	We compare the proposed algorithm with the state-of-the-art methods on multiple real-world benchmarks for unsupervised cross-domain person re-ID in Table~\ref{tab:sota}. Our method consistently outperforms the existing methods by a large margin and achieves the best mAP performance on all the four benchmarks. Specifically, with the same backbone (ResNet-50) and clustering algorithm (DBSCAN), our method surpasses MMT \cite{ge2020mutual} by 8.1\%, 9.9\%, 11.4\% and 11.1\% mAP on the Market-to-Duke, Duke-to-Market, Market-to-MSMT and Duke-to-MSMT benchmarks, respectively. 
	Compared to the other recent unsupervised re-ID methods (e.g., by alternatively training and clustering \cite{yang2019selfsimilarity,zhai2020ad,zhao2020unsupervised} or generating synthetic training data by GAN \cite{chen2019instance}), our PDA method can also obtain superior performance in terms of both mAP and rank-1 accuracy. Compared to the~\cite{ge2020self} which also uses contrastive learning, our method achieves notable gains on the Market-to-MSMT benchmark, e.g., +9.0\% of mAP and +12.1\% of rank-1 accuracy. We implement two supervised baselines with the standard cross-entropy and triplet losses using the ground-truth labels of target data. Our method can achieve similar or comparable results in the challenging unsupervised learning scenario. We also compare with existing unsupervised re-ID methods under the synthetic-to-real setting in Table~\ref{tab:synthetic}. Our method consistently surpasses prior state-of-the-art methods on both benchmarks with a large margin, e.g., outperforming \cite{ge2020self} by 5.8\% of mAP and 6.2\% rank-1 accuracy on PersonX-to-Market. 
	We also evaluate our method on other object re-ID benchmarks~\cite{liu2016deep,liu2016deep2,naphade20204th} in Table \ref{tab:vehicle}. Our method achieves superior performance than MMT and SpCL by a large margin, e.g., +4.3\% and +4.2\% mAP over SpCL on VehicleID-to-VeRi and VehicleX-to-VeRi, repectively.
	

	\begin{table}[t]
		\scriptsize
		\centering
		\begin{center}
			\scalebox{1.0}{
				\begin{tabular}{P{2.8cm}|C{0.7cm}C{0.7cm}|C{0.7cm}C{0.7cm}}
					\toprule
					\multicolumn{1}{c|}{\multirow{2}{*}{Method}} & \multicolumn{2}{c|}{Market-to-Duke} & \multicolumn{2}{c}{Duke-to-Market} \\
					\cmidrule{2-5}
					& mAP & rank-1 & mAP & rank-1 \\ 
					\midrule
					(\ron1). w/o cluster-wise pairs &37.3 &50.7 &46.1 &66.5\\
					(\ron2). w/o past negatives &66.8 &80.0 &78.6 &92.7 \\ 
					(\ron3). w/o instance drop &68.1 &81.4 &81.0 &93.0\\
					(\ron4). $|\mathbb{N}_\text{past}|=512$ &68.5 &82.0 &81.6 &93.4 \\
					(\ron5). $|\mathbb{N}_\text{past}|=1024$ &\textbf{69.4}   &\textbf{82.7} &\textbf{82.2} &\textbf{93.6}\\
					(\ron6). $|\mathbb{N}_\text{past}|=2048$ &67.6 &80.8 &80.7 &92.9 \\
					\bottomrule
				\end{tabular}
			}
		\end{center}
		\vspace{2mm}
		\caption{Ablation studies of the proposed cluster-wise contrastive learning (CCL) algorithm on Market-to-Duke and Duke-to-Market benchmarks. The cross-entropy and triplet losses are used for all the experiments here.}
		\label{tab:ablation-CCL}
		\vspace{-5mm}
	\end{table}
	
	\subsection{Ablation Study}
	\textbf{Contributions from Algorithmic Components.} Table \ref{tab:sota} also shows the relative contributions from each algorithmic component. Our cluster-wise contrastive learning algorithm brings significant improvement over the baseline, e.g., 53.7\% vs. 59.6\% mAP on Market-to-Duke. With the proposed progressive domain adaptation strategy, we obtain another remarkable performance gains, e.g., 59.6\% vs. 67.3\%. By adding the extra loss for training in Fourier space, we can obtain consistent improvements by around 2\% mAP for all the four benchmarks. By combining our method with other training strategies proposed by MMT (e.g., soft losses and mutual learning), we can further obtain improved performance.
	
	\textbf{Ablation studies on contrastive learning.}
	First, we compare the proposed CCL algorithm with three contrastive learning methods in Table~\ref{tab:ablation-CL}. For a fair comparison, we first pre-train the networks on the labeled source data, and then fine-tune these networks with different contrastive losses on target. 
	For the supervised baseline \cite{khosla2020supervised}, we use the GT labels of target data to generate the training pairs. For other unsupervised contrastive learning methods \cite{wu2018unsupervised,he2019momentum}, we find they almost fail on these benchmarks. This is because instance-level supervision is often used to learn general feature representations and its optimization target is different from the re-ID task. Directly applying such instance-wise pairs for training may hinder the discriminability of model to distinguish different high-level classes (i.e., person identities). 
	Second, we conduct detailed ablation experiments on our cluster-wise contrastive learning algorithm in Table~\ref{tab:ablation-CCL}.
	Without cluster-wise pairs (i.e., using instance-wise pairs as in \cite{he2019momentum}), the performance drops significantly on both benchmarks (e.g, only 37.3\% mAP on Market-to-Duke and 46.1\% mAP on Duke-to-Market). Without collecting negative samples in the past iterations (i.e., collecting them from the current batch only) or without dropping positive instances from past features, both experiments obtain degraded performance. We also test different sizes of the queue to store the past features for training and find $|\mathbb(N)_\text{past}|=1024$ performs best in our setting.
	
	\textbf{Effect of progressive weights.}
	We conduct ablation experiments to show the effectiveness of our progressive domain adaptation strategy in Table~\ref{tab:ablation-PDA}. The results of (\ron1) show that the performance is poor by directly testing the pre-trained source model on target without training. This is not surprising since no knowledge is transferred to the unlabeled target domain. Two-stage training (i.e., first pre-training on source and then fine-tuning on target) obtain inferior results compared to our progressive training strategy. We test different combinations of static loss weights for joint training source and target. The best choice ($\lambda_s = 0.2, \lambda_t = 0.8$) is still worse than our progressive weights. We also investigate different multi-step policies as well as the linear policy\footnote{
		linear policy:~$w(e) = \frac{1}{e_1-e_2} \cdot e + \frac{e_2}{e_2-e_1}$
	} and find that 3-step policy performs best in our experiments.

	\begin{table}[t]
		\scriptsize
		\centering
		\begin{center}
			\scalebox{1}{
				\begin{tabular}{P{3cm}|C{0.7cm}C{0.7cm}|C{0.7cm}C{0.7cm}}
					\toprule
					\multicolumn{1}{c|}{\multirow{2}{*}{Method}} & \multicolumn{2}{c|}{Market-to-Duke} & \multicolumn{2}{c}{Duke-to-Market} \\
					\cmidrule{2-5}
					& mAP & rank-1 & mAP & rank-1 \\ 
					\midrule
					(\ron1). source only &31.1 &48.8 &33.7 &62.3 \\
					(\ron2). two-stage training &62.1 &77.7 &74.0 &89.2\\
					(\ron3). static weights (0.8;0.2) &51.9 &67.2 &62.7 &79.1\\
					(\ron4). static weights (0.5;0.5) &56.2 &73.8 &66.8 &84.2\\
					(\ron5). static weights (0.2;0.8) &58.8 &74.9 &68.8 &85.5\\
					(\ron6). 2-step policy &67.8 &81.2 &80.8 &92.4\\
					(\ron7). 3-step policy &\textbf{69.4}   &\textbf{82.7} &\textbf{82.2} &\textbf{93.6} \\
					(\ron8). 4-step policy &68.2 &81.6 &81.1 &92.9 \\
					(\ron9). linear policy &67.6 &81.0 &80.7 &92.5 \\
					
					\bottomrule
				\end{tabular}
			}
		\end{center}
		\vspace{2mm}
		\caption{Ablation studies of the proposed progressive domain adaptation (PDA) strategy on Market-to-Duke and Duke-to-Market benchmarks.}
		\label{tab:ablation-PDA}
		\vspace{-5mm}
	\end{table}
	
	\textbf{Fourier space vs. Spatial space.}
	The motivation of our Fourier augmentation to exploit extra feature space to facilitate the network training. We compare the proposed FA with other alternative nonlinear mappings in Table~\ref{tab:mlp}. The proposed FA outperforms a single MLP layer (FC + ReLU) on both Market-to-Duke and Duke-to-Market benchmarks, which validates the superiority of the proposed method. With the spatial features or the Fourier features only for training, we achieve similar results on both benchmarks (e.g., 67.3\% vs. 67.6\% mAP on Market-to-Duke). By joint training in the spatial and Fourier space, we can further improve the performance. 
	
	\begin{table}[t]
		\scriptsize
		\centering
		\begin{center}
			\scalebox{0.9
			}{
				\begin{tabular}{C{2cm}|C{1cm}C{1cm}|C{1cm}C{1cm}}
					\toprule
					\multicolumn{1}{c|}{\multirow{2}{*}{Method}} & \multicolumn{2}{c|}{Market-to-Duke} & \multicolumn{2}{c}{Duke-to-Market} \\
					\cmidrule{2-5}
					\multicolumn{1}{c|}{}  & mAP & rank-1 & mAP & rank-1 \\ 
					\midrule
					
					\multicolumn{1}{c|}{\multirow{1}{*}{MLP}} &68.2 &81.5  &81.3 &93.2 \\
					\midrule
					\multicolumn{1}{c|}{\multirow{1}{*}{Spatial}} &67.3 &80.9 &80.3 &92.5  \\
					\multicolumn{1}{c|}{\multirow{1}{*}{Fourier}} &67.6 &81.3 &80.8 &92.6 \\
					\multicolumn{1}{c|}{\multirow{1}{*}{Spatial + Fourier}} &\bf69.4 &\bf82.7 &\bf82.2 &\bf93.6\\
					
					\bottomrule
				\end{tabular}
			}
		\end{center}
		\vspace{2mm}
		\caption{Comparisons with different nonlinear mappings.}
		\label{tab:mlp}
		\vspace{-5mm}
	\end{table} 
	
	\begin{table}[t]
		\scriptsize
		\centering
		\begin{center}
			\scalebox{0.85}{
				\begin{tabular}{C{1cm}C{1cm}|C{1cm}C{1cm}|C{1cm}C{1cm}}
					\toprule
					\multicolumn{2}{c|}{Loss weights} & \multicolumn{2}{c|}{Market-to-Duke} & \multicolumn{2}{c}{Duke-to-Market} \\
					\cmidrule{1-6}
					$\delta$ &$\gamma$ & mAP & rank-1 & mAP & rank-1 \\ 
					\midrule
					\multirow{4}{*}{0.01}  &0 &66.7  &80.9 &79.4 &91.2 \\
					&0.5 &67.7  &81.7  &80.1 &92.1  \\
					&0.7 &68.2  &82.4  &80.9 &92.8  \\
					&1  &66.1  &80.2  &78.5 &90.6 \\
					\midrule
					\multirow{4}{*}{0.1}  &0 &67.6 &81.3 &80.8 &92.6 \\
					&0.5 &69.2  &82.4  &81.8 &93.6 \\
					&0.7 &\textbf{69.4}  &\textbf{82.7}  &\textbf{82.2} &\textbf{93.6}  \\
					&1 &67.3  &80.9  &80.3 &92.5   \\
					\midrule
					\multirow{4}{*}{1}  &0  &66.0 &79.8 &77.3 &91.3 \\
					&0.5 &66.8  &80.5  &78.1 &91.9  \\
					&0.7 &67.2  &80.9  &78.7 &92.5  \\
					&1  &65.4  &79.1  &76.5 &90.8  \\ 
					\bottomrule
				\end{tabular}
			}
		\end{center}
		\caption{Ablation studies on loss weights $\delta$ and $\gamma$.}
		\label{tab:lossweight}
		\vspace{-5mm}
	\end{table}
	
	\textbf{Hyper-parameter analysis.}
	To investigate the importance of loss weights $\delta$ and $\gamma$, we conduct experiments by changing $\gamma$ from 0 to 1 under a fixed $\delta$. Table~\ref{tab:lossweight} shows that $\delta=0.1$ and $\gamma=0.7$ perform best on both Market-to-Duke and Duke-to-Market benchmarks. 
	
	\section{Conclusion}
	In this work, we propose a unified framework by incorporating three technical schemes to address the challening unsupervised cross-domain re-ID problem. To learn noise-tolerant feature representations, we propose a cluster-wise contrastive learning algorithm by iterative optimization of feature learning and clustering. Instead of simply fine-tuning the pre-trained source model, we adopt a progressive training mechanism to gradually transfer the knowledge from source to target. Furthermore, We impose extra training constraints on the Fourier space for further maximizing the class separability of re-ID models. Our method consistently outperforms prior unsupervised re-ID methods on multiple benchmarks by a large margin. We believe that an extension of this work is to address large variations (e.g., large poses, partial occlusion) in the unsupervised setting. 
	
	\textbf{Acknowledgement}
	We truly thank Yuwing Tai and Xin Tao for helpful discussion.
	{\small
		\bibliographystyle{ieee_fullname}
		\bibliography{egbib}
	}

	
\end{document}